\documentclass[journal]{IEEEtran}

\usepackage{amsmath,amssymb,amsthm,mathtools}
\usepackage{graphicx}
\usepackage{graphbox}
\usepackage{dblfloatfix}
\usepackage{bbm}
\usepackage[nocompress]{cite}
\usepackage{xcolor}
\usepackage{booktabs}
\usepackage{algorithm}
\usepackage{algorithmic}
\usepackage{comment}
\usepackage{url}

\graphicspath{{figures/}{}}

\newtheorem{proposition}{Proposition}

\DeclareMathOperator*{\LME}{LME}

\newcommand{\centroid}{\boldsymbol{\mu}}
\newcommand{\sv}{\boldsymbol{u}}
\newcommand{\x}{\boldsymbol{x}}
\newcommand{\ba}{\boldsymbol{a}}
\newcommand{\w}{\boldsymbol{w}}
\newcommand{\kmeans}{k-means}
\newcommand{\skdkmeans}{standard\,/\,\allowbreak kernel\,/\,\allowbreak deep \kmeans{}}
\newcommand{\acronym}{NEON}

\usepackage{tikz}
\usepackage[framemethod=tikz]{mdframed}
\newenvironment{boxnetwork}[1][]{%
\mdfsetup{%
	frametitle={%
		\tikz[baseline=(current bounding box.east),outer sep=0pt] \node[anchor=east,rounded corners=3pt,fill=gray!30]
		{\strut\,#1\,};},%
	skipabove=5mm,skipbelow=2mm,%
	leftmargin=0.2cm,rightmargin=0.2cm,
	backgroundcolor=gray!10,
	innertopmargin=-10pt,innerbottommargin=5pt,
	innerleftmargin=2pt,innerrightmargin=2pt,
	linecolor=gray!30,%
	linewidth=1pt,topline=true,%
	roundcorner=3pt,
	frametitlealignment={\hspace*{0.02\linewidth}},
	frametitleaboveskip=-10pt,
	frametitlefont=\normalfont\bf}\begin{mdframed}[nobreak=true]\relax%
}{\end{mdframed}}

\begin{document}

\title{From Clustering to Cluster Explanations\\via Neural Networks}

\author{Jacob Kauffmann, Malte Esders, Lukas Ruff, Gr\'egoire Montavon${^*}$, Wojciech Samek, Klaus-Robert M\"uller${^*}$
\thanks{J. Kauffmann is with the Berlin Institute of Technology (TU Berlin), 10587 Berlin, Germany.}
\thanks{M. Esders is with the Berlin Institute of Technology (TU Berlin), 10587 Berlin, Germany.}
\thanks{L. Ruff is with Aignostics, 10117 Berlin, Germany.}
\thanks{G. Montavon is with the Berlin Institute of Technology (TU Berlin), 10587 Berlin, Germany; and BIFOLD -- Berlin Institute for the Foundations of Learning and Data, 10587 Berlin, Germany. E-mail: gregoire.montavon@tu-berlin.de.}
\thanks{W. Samek is with Fraunhofer Heinrich Hertz Institute, 10587 Berlin, Germany; and BIFOLD -- Berlin Institute for the Foundations of Learning and Data, 10587 Berlin, Germany. E-mail: wojciech.samek@hhi.fraunhofer.de }
\thanks{K.-R. M\"uller is with the Berlin Institute of Technology (TU Berlin), 10587 Berlin, Germany; BIFOLD -- Berlin Institute for the Foundations of Learning and Data, 10587 Berlin, Germany; the Department of Artificial Intelligence, Korea University, Seoul 136-713, Korea; and the Max Planck Institut f{\"u}r Informatik, 66123 Saarbr{\"u}cken, Germany. E-mail: klaus-robert.mueller@tu-berlin.de.}
\thanks{(Corresponding Authors marked with asterisk: Gr\'egoire Montavon, Klaus-Robert M\"uller)}}

\maketitle

\begin{abstract}
A recent trend in machine learning has been to enrich learned models with the ability to explain their own predictions. The emerging field of Explainable AI (XAI) has so far mainly focused on supervised learning, in particular, deep neural network classifiers. In many practical problems however, label information is not given and the goal is instead to discover the underlying structure of the data, for example, its clusters. While powerful methods exist for extracting the cluster structure in data, they typically do not answer the question \emph{why} a certain data point has been assigned to a given cluster. We propose a new framework that can, for the first time, explain cluster assignments in terms of input features in an efficient and reliable manner. It is based on the novel insight that clustering models can be rewritten as neural networks---or `neuralized'. Cluster predictions of the obtained networks can then be quickly and accurately attributed to the input features. Several showcases demonstrate the ability of our method to assess the quality of learned clusters and to extract novel insights from the analyzed data and representations.
\end{abstract}


\section{Introduction}

Clustering is an important class of unsupervised learning models that aims to reflect the intrinsic heterogeneities of common data generation processes \cite{DBLP:journals/csur/JainMF99,DBLP:journals/tnn/XuW05,Jain:1988:ACD:46712, Hastie2009}. Natural cluster structures are observed in a variety of contexts from e.g.~gene expression \cite{DBLP:journals/tkde/JiangTZ04} and ecosystems composition \cite{Celiker2014} to textual data \cite{DBLP:conf/emnlp/MekalaGPK17}. Methods that can accurately identify the cluster structure have thus been the object of sustained research over the past decades \cite{DBLP:journals/prl/Jain10}. Basic techniques such as \kmeans{} \cite{macqueen1967} have been extended to operate in kernel feature spaces \cite{DBLP:journals/pami/ShiM00, DBLP:conf/kdd/DhillonGK04}, or on the representations built by a deep neural network \cite{DBLP:conf/icml/XieGF16,DBLP:conf/icassp/HersheyCRW16,DBLP:conf/icml/YangFSH17,DBLP:conf/eccv/CaronBJD18}.

Due to the ever growing complexity of ML models and their use in increasingly sensitive applications, it has become crucial to endow these models with the capability to explain their own predictions in a way that is interpretable for a human. Explainable AI (XAI) has emerged as an important direction for machine learning, and excellent results have been reported in selected tasks such as explaining the predictions of popular DNN classifiers \cite{DBLP:conf/eccv/ZeilerF14, Bach2015, DBLP:conf/kdd/Ribeiro0G16, DBLP:conf/iccv/SelvarajuCDVPB17,samek2019explainable}.

\smallskip

In this paper, we bring these newly developed explanation capabilities to clustering, a highly needed functionality, considering that in the first place one of the main motivations for performing a clustering is knowledge discovery. Especially in high-dimensional feature space, a clustering for knowledge discovery can only provide a few prototypical data points for each cluster. Such prototypes, however, do not reveal which features made them prototypical. Instead, we would like to let the clustering model explain itself in terms of the very features that have contributed to the cluster assignments.---To the best of our knowledge, our work is the first ever attempt to systematically and comprehensively obtain such explanations. Specifically, we are able to supply explanations of {\em why} each individual point is clustered in the way it is.

The method we propose, puts forward the novel insight that a broad range of clustering models can be rewritten, \textit{without retraining}, as \textit{functionally equivalent} neural networks, which then serve as a backbone to guide the explanation process. Technically, we suggest to apply the following two steps: (1) The cluster model is `{\em neuralized}' by rewriting it as a functionally equivalent neural network with standard detection/pooling layers. (2) Cluster assignments formed at the output of the neural network are then {\em propagated} backwards using an LRP-type procedure (cf.~\cite{Bach2015,DBLP:journals/dsp/MontavonSM18,samek2021explaining}) until the input variables (e.g.\ pixels or words) are reached.
 
\begin{figure*}[t!]
\centering
\includegraphics[width=.975\linewidth]{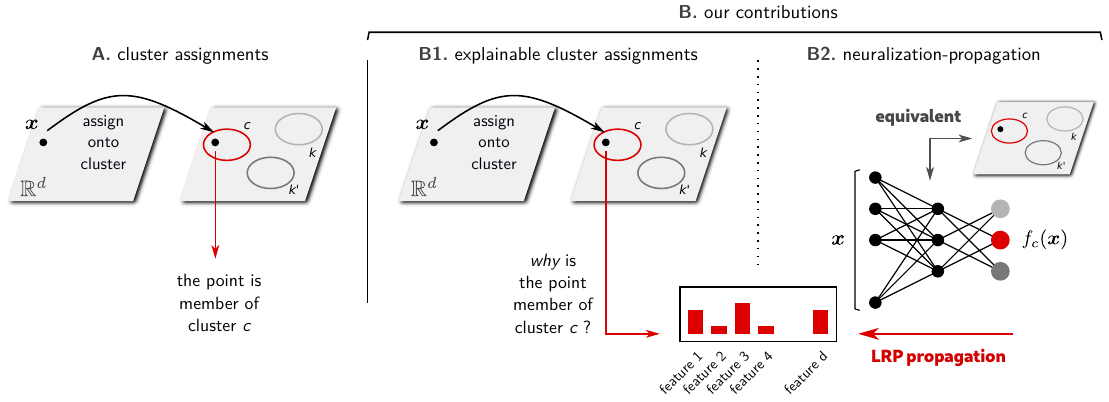}
\caption{From clustering to cluster explanations via neural networks. \textbf{A.} Standard clustering scenario where data are assigned onto clusters according to the clustering model. \textbf{B.} Overview of our contributions. \textbf{B1.} We enrich the cluster assignment with an explanation highlighting what input features mostly contribute to the cluster decision. \textbf{B2.} We achieve this technically by observing that the clustering decision can be rewritten as a neural network (neuralization), enabling fast and robust explanations via the LRP technique (propagation).}
\label{fig:cartoon}
\end{figure*}

The proposed `neuralization-propagation' procedure (or short, \acronym{}) is tested on a number of datasets and clustering models, including recent deep clustering models such as SCAN \cite{DBLP:conf/eccv/GansbekeVGPG20}. Each time, NEON accurately explains cluster assignments, and extracts useful insights. Experiments also demonstrate the practical value of our two-step approach compared to a potentially simpler one-step approach without neuralization. Our contributions can be summarized as follows:
\begin{itemize}
    \item Introduction of XAI to clustering, specifically, explanation of the assignment of individual data points onto clusters, in terms of input features.
    \item Formulation of the clustering decisions for a broad range of clustering models as being functionally equivalent neural networks, thus enabling the application of state-of-the-art XAI techniques to these models.
    \item Theoretical embedding of our neuralization-propagation approach to explaining clustering, specifically providing an interpretation of our approach, for special cases, in terms of Shapley Values.
    \item Demonstration of the benefit of bringing XAI to clustering showcased for two real-world examples, and extensive quantitative validation of our proposed explanation method.
\end{itemize}

Fig.\ \ref{fig:cartoon} shows a cartoon of our contributions in order to provide the general underlying intuition to the reader.
We stress that our method applies to many popular clustering algorithms and is a generic blueprint as it does neither rely on predesigned interpretability structures nor algorithms, nor any retraining. This will prove useful in the future for shedding new light into existing cluster-based typologies used e.g.\ in computational biology \cite{Tavazoie1999,Ciriello2013} or consumer data \cite{KengKau2003}, which researchers and practitioners have started to use increasingly to support their scientific reasoning and to take insightful decisions.

\subsection{Related Work}

So far, research on explanation methods has been overwhelmingly focused on the case of supervised learning. Methods based on the gradient \cite{DBLP:conf/iscas/ZuradaMC94, DBLP:journals/jmlr/BaehrensSHKHM10, DBLP:conf/icml/SundararajanTY17}, local perturbations \cite{DBLP:conf/eccv/ZeilerF14, DBLP:conf/nips/LundbergL17}, or surrogate functions \cite{DBLP:conf/kdd/Ribeiro0G16} do not make specific assumptions about the structure of the model and are thus applicable to a wide range of classifiers. Other methods require the classifier to have a neural network structure and apply a purposely designed backward propagation pass \cite{Bach2015, DBLP:journals/pr/MontavonLBSM17, DBLP:conf/cidm/LandeckerTBMKB13, DBLP:conf/icml/ShrikumarGK17,DBLP:journals/dsp/MontavonSM18,samek2021explaining} to produce accurate explanations at low computational cost. While recent work has extended the principle to other types of models such as one-class SVMs \cite{DBLP:journals/pr/KauffmannMM20}, LSTM networks \cite{DBLP:conf/wassa/ArrasMMS17}, or graph neural networks \cite{schnake2021higher}, the method we propose here  contributes by offering a solution to the so far unsolved problem of explaining cluster assignments. 

Note that the few cluster interpretability techniques have so far been based on surrogate decision trees \cite{schafer2005annealed,DBLP:journals/corr/abs-1812-00539, DBLP:journals/adac/FraimanGS13, DBLP:conf/icml/MoshkovitzDRF20, DBLP:journals/bmcbi/GeurtsTDd07}, where the decision tree is trained to approximate the \kmeans{} clustering as closely as possible, and where the cluster assignment is then interpreted using Explainable AI techniques specific to decision trees. With such a surrogate approach, the user typically has to trade off faithfulness to the original model against explainability.

Related to the connections we establish in this paper between clustering models and neural networks, some works explore ways of merging the two in order to produce better, more flexible ML models. For example deep clustering approaches typically build a clustering objective on top of deep representations \cite{DBLP:conf/icml/YangFSH17, DBLP:conf/eccv/CaronBJD18, DBLP:conf/iccv/DizajiHDCH17, DBLP:conf/icml/XieGF16, DBLP:conf/iconip/GuoLZY17}. Other models, in particular, the $k$-meansNet \cite{DBLP:journals/corr/abs-1808-07292} design the neural network in a way that simulates a clustering model, so that the learned neural networks solution can be interpreted as a clustering solution. Note that in all these works, the purpose is more to enhance a basic clustering model by providing the flexibility of neural network representation and training, whereas our work focuses on making existing popular clustering algorithms explainable.

Another set of related works focus on the problem of {\em learning} a good clustering model, by identifying a subset of relevant features that support the cluster structure. Some methods identify relevant features by running the same clustering algorithm multiple times on different feature subsets \cite{DBLP:journals/jmlr/DyB04}. Other approaches simultaneously solve feature selection and clustering by defining a joint objective function to be minimized \cite{DBLP:journals/pami/LawFJ04}. While feature selection can identify the set of features required to represent the overall cluster structure, our work builds up by identifying among those features which ones are truly responsible for a given cluster or a given cluster assignment.

Further related works focus on quantitatively validating clustering solutions. Examples of validation metrics are compactness\,/\,separation of clusters \cite{DBLP:journals/jiis/HalkidiBV01}, cluster stability under resampling\,/\,perturbations \cite{DBLP:journals/neco/LangeRBB04,DBLP:conf/nips/Meila18}, or purity, i.e.\ the absence of examples with different labels in the same cluster \cite{DBLP:books/daglib/0021593}. Our work enhances the validation of clustering models by producing human-interpretable feedback, a critical step to identify whether cluster assignments are supported by meaningful features or by what the user would consider to be artifacts.

Lastly, user interfaces have been developed to better navigate cluster structures, as they occur, e.g.\ in biology applications \cite{Metsalu2015,Kern2017}. Also, the use of prototypes has been proposed to visualize deep image clustering models \cite{DBLP:conf/eccv/CaronBJD18} or explain kernel methods for property prediction of chemical compounds \cite{hansen2011visual}. While these works produce useful and informative visualizations which may help to guide the process of clustering, our approach contributes by answering the precise question ``{\em why} a given data point is assigned to a particular cluster.''

\section{Explaining K-Means Cluster Assignments}
\label{section:standard}

The \kmeans{} algorithm \cite{macqueen1967} is one of the best known approaches to clustering and is used in many scientific and industrial applications (e.g.\ \cite{Dhaeseleer2005,DBLP:conf/www/Sculley10,Hanson2020}). This section presents our neuralization-propagation approach for explaining a k-means cluster assignment in terms of input features. Due to the simplicity of the k-means model, this section also has a tutorial purpose. More complex and powerful clustering models based on kernels \cite{DBLP:conf/kdd/DhillonGK04}, deep neural networks \cite{DBLP:conf/icml/XieGF16,DBLP:conf/icml/YangFSH17}, or more general clustering techniques, are discussed in Sections \ref{section:kernel}--\ref{section:any}.

The k-means algorithm finds a set of centroids that minimizes the total squared distance between each data point and their nearest centroid. The k-means model assigns points to clusters based on their distance to each centroid $\centroid_k\in\mathbb{R}^d$, specifically the model assigns a point $\x \in\mathbb{R}^d$ to cluster $c$ if
\begin{align}
\forall_{k \neq c}:~ \|\x-\centroid_c\|^2 < \|\x-\centroid_k\|^2 .
\label{eq:decision}
\end{align}
In principle, it is conceivable to use Explainable AI techniques such as prediction difference analysis \cite{DBLP:conf/eccv/ZeilerF14,DBLP:conf/iclr/ZintgrafCAW17} or LIME \cite{DBLP:conf/kdd/Ribeiro0G16}, as they apply out-of-the-box to \textit{any} model or decision function.
However, these approaches require to evaluate the function multiple times to test for the effect of each input dimension. This can become slow when the data is high-dimensional, e.g.\ when clustering images or gene expression data \cite{Ciriello2013}. Also, local perturbation may not faithfully depict the overall contribution of a feature to the clustering decision, especially if multiple features needs to be perturbed in order to affect the decision.

\smallskip

In the context of supervised learning, more efficient Explainable AI techniques have been proposed, which rely on a model that induces the decision function, and from which meaningful gradient information and intermediate representations can be extracted. Such methods include, among others, integrated gradients \cite{DBLP:conf/icml/SundararajanTY17}, or Layer-wise Relevance Propagation (LRP) \cite{Bach2015,DBLP:series/lncs/MontavonBLSM19,samek2021explaining}. The LRP method in particular, leverages the neural network structure of the prediction to produce a robust explanation in the order of a single forward/backward pass. The LRP method was used in a wide range of applications (e.g.\ \cite{LapICCVW17,DBLP:conf/acl/DingLLS17,Horst2019, DBLP:journals/jstsp/PerotinSVG19,samek2021explaining,Eberle2020,anders2022finding,schnake2021higher}), and can be embedded in the framework of deep Taylor decomposition \cite{DBLP:journals/pr/MontavonLBSM17}.

\subsection{Neuralization of the Cluster Assignment}
\label{section:standard-neuralization}

In order to bring these efficient XAI techniques to clustering, we propose to enrich the clustering decision function $g_c(\x)$ with a neural network model. The latter is designed to exactly replicate the cluster assignments of the original clustering model and is more amenable to explainability. Furthermore, we also require that such neural network model is obtained readily from the cluster solution (i.e.\ the centroids) without incurring any additional training step. We call the process of obtaining such a neural network ``neuralization.''

\begin{figure}[h]
\centering
\includegraphics[width=.35\linewidth]{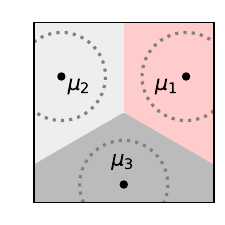}\hskip 3mm
\includegraphics[width=.35\linewidth]{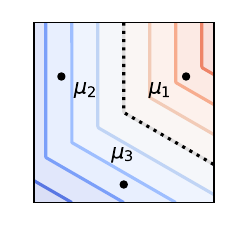}
\caption{Left: Decision function of a k-means clustering model with centroids $\centroid_1,\centroid_2,\centroid_3$. Data points in the region highlighted in red are assigned to the cluster $c=1$. Right: Contour plot of the function $f_c(\x)$ for the cluster $c=1$.}
\label{fig:neuralization}
\end{figure}

\begin{proposition}
The decision function of Eq.\ \eqref{eq:decision} can be reproduced by a two-layer neural network composed of a standard linear layer and a (min-)pooling layer:
\begin{boxnetwork}[Neuralized $\boldsymbol k$-means]
\begin{align*}
h_k &= \w_k^\top \x + b_k & (\mathrm{layer}~1)\\
f_c &= \min_{k \neq c} \{h_k\}& (\mathrm{layer}~2)
\end{align*}
\end{boxnetwork}
where $\w_k = 2 (\centroid_c - \centroid_k)$ and $b_k = \|\centroid_k\|^2 - \|\centroid_c\|^2$, and assigning to cluster $c$ if $f_c(\x) > 0$.
\label{proposition:neuralization}
\end{proposition}
\noindent (cf.\ Appendix A of the Supplement for a derivation).
The first layer corresponds to a collection of linear functions aligned with the different cluster centroids. The min-pooling selects which linear function is active at a given location. These two layers together build a piecewise linear function. A simple two-dimensional example with three clusters is shown in Fig.\ \ref{fig:neuralization}. We observe that the neural network output $f_c(\x)$ (right) exactly reproduces the true cluster decision boundary, specifically, the Voronoi partition associated to the given k-means model (left).

The neural network above can be also interpreted in neuroscientific terms as the alternation of `simple cells' and `complex cells' \cite{Hubel1962}, or `executive organs' and `restoring organs' in automata theory \cite{von1956probabilistic}. We also note that earlier works have already linearized elements of the cluster model such as the square distance for the purpose of training \cite{DBLP:journals/corr/abs-1808-07292}. Here, our contribution differs by extracting a piecewise linear view of the \textit{whole} model, and additionally, identifying a neural network structure for this piecewise linear form. We provide similar neuralization results for the soft k-means case, as well as a probabilistic interpretation, in Appendix E of the Supplement. We will also study more complex neuralization scenarios in Sections \ref{section:kernel} and \ref{section:deep} when considering kernel-based clustering and deep clustering.

\subsection{Propagation of the Cluster Assignment}

So far, we have rewritten the k-means decision function for each cluster as a neural network. This initial step gives access to a broader range of explanation techniques such as integrated gradients \cite{DBLP:conf/icml/SundararajanTY17}, or layer-wise relevance propagation (LRP) \cite{Bach2015,DBLP:series/lncs/Montavon19}. The LRP technique, in particular, leverages the neural network structure to produce robust explanation in a single forward/backward pass. Unlike the standard gradient propagation pass which provides a highly localized view of the function, LRP applies propagation rules that redistribute the quantity to explain from layer to layer. These rules are purposely designed for the task of explanation. LRP ensures certain desirable properties of an explanation such as conservation of predicted evidence and local continuity of explanations \cite{Bach2015,DBLP:series/lncs/Montavon19}.

\medskip

Let us start with the output of the neural network $f_c$, which we wish to attribute as a first step to neurons in the intermediate layer $(h_k)_k$, by propagating through the min function. Similar to \cite{DBLP:journals/pr/KauffmannMM20}, we follow a min-take-most (MTM) strategy, where smallest inputs to that function receive the largest share of the quantity to redistribute, in particular, we apply the propagation rule:
\begin{align}
R_k = \frac{\exp(-\beta h_k)}{\sum_{k \neq c} \exp(-\beta h_k)} f_c
\label{eq:mintakemost}
\end{align}
where $R_k$ is the `relevance' of neuron $h_k$ to the cluster assignment $f_c$, and where $\beta\in\mathbb{R}^+$ is a stiffness hyperparameter. The stiffness parameter interpolates between a uniform redistribution strategy ($\beta = 0$) and a min-take-all strategy ($\beta \to \infty$). 
Note that compared to these two extreme cases, our approach allows to contextualize the explanation (i.e.\ not redistributing on clusters competitors that are too far and therefore irrelevant), and at the same time, ensures continuity of the explanation as we transition from one nearest cluster competitor to another. We propose to set this parameter according to the simple heuristic:
\begin{align}
\beta = {\mathbb E}[f_c]^{-1}
\label{eq:heuristic}
\end{align}
where the expectation is computed over the whole dataset. In other words, considering $f_c$ to be a `typical' score in the pool, we want the stiffness parameter to be inversely proportional to it.

\medskip

We now consider how to further redistribute the intermediate relevance scores $R_k$ to the input layer, where the dimensions correspond to observed quantities that are assumed to be interpretable by the user. To achieve this, we propose the LRP propagation rule:
\begin{align}
R_i = \sum_{k \neq c} \frac{(x_i - m_{ik})\cdot w_{ik}}{\sum_i (x_i - m_{ik})\cdot w_{ik}} R_k
\label{eq:directional}
\end{align}
where $\boldsymbol{m}_{k} = (\centroid_c + \centroid_k)/2$ is the mid-point between the centroids of the cluster of interest and the competitor. In other words, we attribute on dimensions where the input activation relative to the mid-point, $\x - \boldsymbol{m}_k$, matches the model response $\w_k$.

It can be noted that the proposed propagation rules ensure a certain number of desirable properties of an explanation, in particular, it satisfies the conservation property $\sum_i R_i = f_c(\x)$, it preserves the continuity of $f_c(\x)$, and it is invariant to any translation of the clustering in input space.

\subsection{Theoretical Embedding}

We provide further theoretical support for the rules in Eqs.\ \eqref{eq:mintakemost} and \eqref{eq:directional} by showing that their application produce, for special cases, explanations that coincide with the Shapley Value. The Shapley Value \cite{Shapley,DBLP:journals/jmlr/StrumbeljK10,DBLP:conf/nips/LundbergL17}, originally proposed in the context of game theory, is an axiomatic solution to the problem of attributing the value of a coalition of players to individual players in the coalition.
For our comparison, we interpret the set of players as the individual input features (or activations) and the withdrawal of a player from the coalition as replacing the corresponding feature value $x_i$ by some reference value $\widetilde{x}_i$.

\begin{proposition}
Redistribution performed by Eq.\ \eqref{eq:mintakemost} with parameter $\beta = 0$, corresponds to the Shapley Value of the function $f_c(\boldsymbol{h})$ with the reference point $\widetilde{\boldsymbol{h}} = \boldsymbol{0}$.
\label{proposition:shapley-hidden}
\end{proposition}

(The proof is given in Appendix B of the Supplement.) The parameter $\beta=0$ corresponds to a uniform redistribution of $f_c$ to the cluster competitors. The corresponding reference point $\widetilde{\boldsymbol{h}} = \boldsymbol{0}$ can be interpreted as the image of a point $\widetilde{\x}$ in input space that is equidistant from all cluster centroids. (Note that this point may not exist in low-dimensional spaces.)

\begin{proposition}
When the number of clusters is equal to $2$, the model reduces to $f_c(\x) = \w_k^\top \x + b_k$, and redistribution by Eqs.\ \eqref{eq:mintakemost} and \eqref{eq:directional} corresponds to the Shapley Value of the function $f_c(\x)$ with the reference point $\widetilde{\x} = \boldsymbol{m}_k$.
\label{proposition:shapley-twoclusters}
\end{proposition}

(See Appendix B of the Supplement for a proof.) In other words, the explanation coincides with Shapley values with the reference point $\widetilde{\x}$ chosen at the mid-point between the clusters centroids $\centroid_k$ and $\centroid_c$. Such reference point is a natural choice for explaining why a point is member of cluster $c$ and not of cluster $k$.

\section{Extension to Kernel K-Means}
\label{section:kernel}

The standard k-means clustering algorithm has strong limitations in terms of representation power, as it only allows to represent clusters that are pairwise linearly separable. The kernel k-means model \cite{DBLP:conf/kdd/DhillonGK04} is a straightforward extension of k-means where the data is first mapped to a feature space via some map $\x \mapsto \Phi(\x)$ induced by some kernel function $\mathbb{K}(\x,\sv)$. The decision function implemented by kernel k-means is given by:
\begin{align}
\forall_{k \neq c}:~ \|\Phi(\x)&-\centroid_c\|^2\nonumber \\[1mm]
&< \|\Phi(\x)-\centroid_k\|^2
\label{eq:kernel-decision}
\end{align}
where the centroids are also defined in feature space.

\smallskip

If we were to apply the same explanation framework as in Section \ref{section:standard}, we would obtain an explanation in terms of dimensions of the feature space, and we would then need to further backpropagate through the feature map $\Phi$. While this is technically possible (e.g.\ for a Gaussian kernel $\mathbb{K}(\x,\sv) = \exp(-\gamma \|\x-\sv\|^2)$, one can use random approximations of the feature map), we consider instead a more intuitive formulation, specific to the Gaussian kernel case, where the distance to a particular cluster is modeled by a soft minimum over distance to the cluster members. Specifically, we consider in place of Eq.\ \eqref{eq:kernel-decision} the decision function
\begin{align}
\forall_{k \neq c}:~ {\LME_{i \in \mathcal{C}_c}}^{-\gamma} &\big\{\|\x - \sv_i\|^2\big\}\nonumber\\
&< {\LME_{j \in \mathcal{C}_k}}^{-\gamma} \big\{\|\x - \sv_j\|^2\big\}
\label{eq:kernel-decision-improved}
\end{align}
where $(\sv_i)_{i}$ and $(\sv_j)_{j}$ are sets of data points (or support vectors) representing the two clusters, ${\cal C}_c, {\cal C}_k\subset\mathbb{N}$ are the non-overlapping sets of indices of support vectors that represent these clusters, and where $\LME^{-\gamma}$ denotes a generalized $F$-mean with $F(t) = e^{-\gamma t}$, i.e.\
\begin{align}
{\LME_{i \in \mathcal{C}}}^{-\gamma}\{s_i\} = -\frac{1}{\gamma} \log \Big( \frac{1}{|\mathcal{C}|} \sum_{i \in \mathcal{C}} \exp(-\gamma s_i)\Big).
\end{align}
The latter can be interpreted as a soft min-pooling and it converges to a hard min-pooling when $\gamma \to \infty$.

\smallskip

\begin{figure}[h]
\centering
\parbox{.45\linewidth}{\centering \small ~~$\|\Phi(\x) - \centroid_c\|^2$}
\parbox{.45\linewidth}{\centering \small ~~$\LME_{i \in \mathcal{C}_c}^{-\gamma}\{\cdot\}$}\\
\includegraphics[width=.45\linewidth]{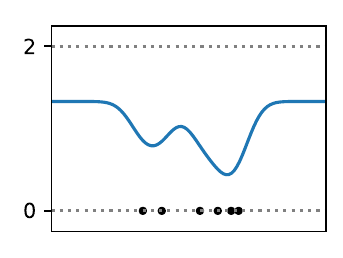}
\includegraphics[width=.45\linewidth]{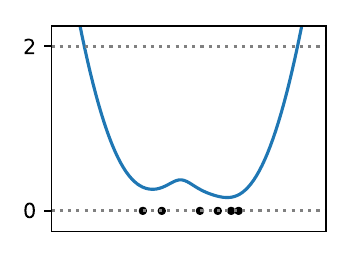}
\caption{Distance between some data point $\x$ and a cluster $c$ depicted as a collection of black dots. The distance is either computed in feature space, or using the soft min-pooling of Eq.\ \eqref{eq:kernel-decision-improved}.}
\label{fig:kernel1d}
\end{figure}

The two distance measures on which the decision functions of Eqs.\ \eqref{eq:kernel-decision} and \eqref{eq:kernel-decision-improved} are based, are illustrated for some toy one-dimensional cluster $c$ composed of 6 data points in Fig.~\ref{fig:kernel1d}.

While the two functions clearly differ, one can also observe that they build comparable level sets. In fact, we show in Proposition \ref{proposition:kernel-approximation} that these two measures of distance are essentially the same up to some monotonous nonlinear transformation, thereby leading to the same decision function.

\begin{proposition}
Let $\centroid_c = \frac{1}{Z_c} \sum_{i \in \mathcal{C}_c} \Phi(\sv_i)$ where $\Phi$ is some feature map associated to the Gaussian kernel $\mathbb{K}(\x,\sv) = \exp(-\gamma \|\x-\sv\|^2)$ and $Z_c$ is a normalization factor. The two distance functions appearing in Eqs.\ \eqref{eq:kernel-decision} and \eqref{eq:kernel-decision-improved} can be related as:
\begin{align}
{\LME_{i \in \mathcal{C}_c}}^{-\gamma} \big\{\|\x - \sv_i\|^2\big\} = g_c(\|\Phi(\x) - \centroid_c\|^2)
\end{align}
where $g_c$ is a monotonically increasing function defined as:
\begin{align}
g_c(\xi) &= \gamma^{-1} \mathrm{Li}_1(\xi/2+\Delta_c) + \gamma^{-1} H_c
\end{align}
with $\mathrm{Li}_1$ is the polylogarithm of order $1$, $\Delta_c = (1 - \|\centroid_c\|^2) / 2$, and $H_c = \log(|\mathcal{C}_c| / Z_c)$.
\label{proposition:kernel-approximation}
\end{proposition}

A proof is given in Appendix C of the Supplement. Formally, equivalence between the two decision functions (Eqs.\ \eqref{eq:kernel-decision} and \eqref{eq:kernel-decision-improved}) is ensured when the function $g_c$ does not depend on the choice of cluster $c$. When choosing the normalization factor $Z_c=|\mathcal{C}_c|$ (standard kernel k-means), the term $H_c$ vanishes but the term $\Delta_c$ remains, and the converse happens if setting $\|\centroid_c\|=1$, i.e.\ $Z_c = \|\sum_{i \in \mathcal{C}_c} \Phi(\sv_i)\|$ (spherical kernel k-means). In practice, both terms remain near zero if we observe that each cluster is equally heterogeneous and has consequently the same norm in feature space. In that case, the two decision boundaries become equivalent. An advantage of the latter decision function is that it can exactly reproduced by a neural network.

\begin{proposition}
The decision function in Eq.\ \eqref{eq:kernel-decision-improved} can be reproduced by a four-layer neural network composed of a linear layer followed by three pooling layers:
\begin{boxnetwork}[Neuralized kernel \kmeans{}]
\begin{align*}
h_{ijk} &= \w_{ij}^\top \x + b_{ij} & (\mathrm{layer}~1)\\
h_{jk} &= {\LME_{i \in \mathcal{C}_c}}^{\gamma} \{ h_{ijk} \} & (\mathrm{layer}~2)\\
h_{k} &= {\LME_{j \in \mathcal{C}_k}}^{-\gamma} \{ h_{jk} \} & (\mathrm{layer}~3)\\
f_c &= \min_{k \neq c} \{h_k\} & (\mathrm{layer}~4)
\end{align*}
\end{boxnetwork}
where $\w_{ij} = 2 \cdot (\sv_i - \sv_j)$ and $b_{ij} = \|\sv_j\|^2 - \|\sv_i\|^2$,
where $\LME^\gamma$ and $\LME^{-\gamma}$ can be interpreted as soft max-pooling and soft min-pooling respectively, and assigning to cluster $c$ if $f_c(\x)>0$.
\label{proposition:kernel-neuralization}
\end{proposition}

\begin{figure}[h]
\centering
\includegraphics[width=.4\linewidth]{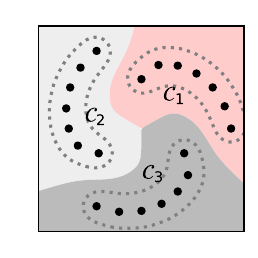}
\includegraphics[width=.4\linewidth]{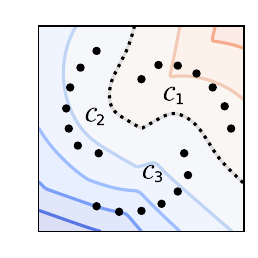}
\caption{Left: Partition implemented by a kernel k-means clustering with three clusters supported by seven support vectors each. Right: Neural network output $f_c(\x)$ associated to the first cluster.}
\label{fig:kernel}
\end{figure}

The proof is given in Appendix D of the Supplement. An example showing the equivalence between the neural network output and Eq.\ \eqref{eq:kernel-decision-improved} is given in Fig.\ \ref{fig:kernel}.

This neural network we have proposed can now be used to support the process of explanation. Because the network is again composed of linear and pooling layers, propagation rules proposed for the k-means case remain applicable. In particular, redistribution in pooling layers can be achieved using Eq.\ \eqref{eq:mintakemost} (and switching the sign for the soft max-pooling case).\footnote{The relevance attributed to neuron $h_{ijk}$ is thus given as $$R_{ijk} = \frac{\exp(\gamma h_{ijk})}{\sum_{i\in{\cal C}_c}\exp(\gamma h_{ijk})}\cdot\frac{\exp(-\gamma h_{jk})}{\sum_{j\in{\cal C}_k}\exp(-\gamma h_{jk})}\cdot R_k.$$} The directional redistribution in the first layer can be achieved using Eq.\ \eqref{eq:directional}. However, we must handle the case where some relevance lands on a deactivated (or weakly activated) neuron $h_{ijk}$, as the latter does not provide directionality in input space. Such special case can be handled by only propagating part of the relevance (and dissipating the rest), specifically, by performing the reassignment:
\begin{align}
R_{ijk} \gets R_{ijk} \cdot (h_{ijk} / h_k) 
\end{align}
The latter ensures that the relevance continuously converges to zero as the neuron $h_{ijk}$ becomes deactivated.

\smallskip

In terms of computational cost, we note that the number of neurons in our neuralized k-means model grows quadratically with the number of support vectors per cluster, whereas the complexity of a simple evaluation of the decision function is linear with the number of support vectors. For NEON to remain computationally favorable, the number of support vectors must be kept small, typically, in the order of $10$ support vectors per cluster. Practical approaches to produce a limited number of support vectors include e.g.\ reduced sets \cite{DBLP:journals/tnn/ScholkopfMBKMRS99,muller2001introduction,DBLP:books/lib/ScholkopfS02}, vector quantization \cite{DBLP:conf/icpr/ZhangR02}, or representing each cluster as a mixture model with finitely many mixture elements (we use this approach in Section \ref{section:text-explanation}). Alternatively, when for modeling purposes it is necessary to maintain a large number of support vectors per cluster, one can adopt a pruning strategy, where we only evaluate in the forward and backward pass the most relevant part of the network, i.e.\ neurons to which the min- and max-pooling functions in the network are effectively sensitive.

\begin{figure*}
\centering
\includegraphics[width=\textwidth]{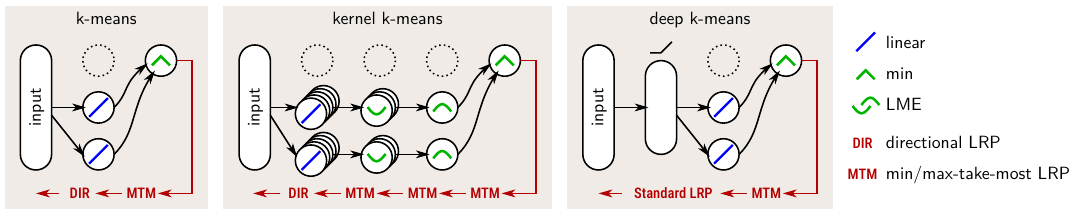}
\caption{Examples of clustering models whose cluster assignments can be explained with our NEON approach. The neuralized models, each of which can be expressed as combinations of detection layers and pooling layers, are depicted along with the propagation rules applied at each layer.}
\label{figure:networks}
\end{figure*}

\section{Extension to Deep Clustering}
\label{section:deep}

Unlike kernel k-means, deep k-means makes use of a feature map given explicitly as a sequence of layer-wise mappings $\Psi(\x) = \Psi_L \circ \dots \circ \Psi_1(\x)$, and the feature map is typically learned via backpropagation to produce the desired cluster structure.

Various formulations of deep k-means have been proposed in the literature. Clustering solutions produced by \cite{DBLP:conf/icml/YangFSH17,DBLP:conf/eccv/CaronBJD18} optimize a hard \kmeans{} objective based on distances in feature space. Using the same assignment model as for k-means, but this time in feature space, we decide for cluster $c$ if:
\begin{align}
\forall_{k \neq c}:~\|\Psi(\x)&-\centroid_c\|^2 \nonumber\\[1mm]
&< \|\Psi(\x)-\centroid_k\|^2
\label{eq:deep-discriminant}
\end{align}
This lets us rewrite the full model as a the stacking of the $L$ layers of the neural network $\Psi$ with the neuralized $k$-means model defined in Proposition \ref{proposition:neuralization}:

\medskip
\begin{boxnetwork}[Neuralized deep $\boldsymbol k$-means]
\begin{align*}
\ba &= \Psi_L \circ \dots \circ \Psi_1(\x) & \text{(layers $1\dots L$)}\\
h_k &= \w_k^\top \ba + b_k & \text{(layer $L+1$)}\\
f_c &=  \min_{k \neq c} \{h_k\} & \text{(layer $L+2$)}
\end{align*}
\end{boxnetwork}
where $\w_k = 2 \cdot (\centroid_c-\centroid_k)$ and $b_k = \|\centroid_k\|^2 - \|\centroid_c\|^2$. Note that beyond a simple application of standard k-means on top of a given layer, there have been many proposals for deep clustering. 

\smallskip

Other quite popular formulations make use of a soft cluster assignment model, specifically, a softargmax model \cite{DBLP:conf/iccv/DizajiHDCH17,DBLP:conf/eccv/GansbekeVGPG20}, or a t-Student similarity model \cite{DBLP:conf/icml/XieGF16,DBLP:conf/iconip/GuoLZY17}. These soft clustering approaches bring a probabilistic interpretation of cluster assignments, and enable entropy-based optimization criteria. In the soft k-means models of \cite{DBLP:conf/iccv/DizajiHDCH17,DBLP:conf/eccv/GansbekeVGPG20}, the data is first projected on some direction $\centroid_c$ associated to the cluster, and mapped to a probability score using a softmax. Here we first consider the explanation of the clustering outcome, in other words, we place the decision boundary at the location where there is as much evidence for the given cluster assignment as for the assignment onto the nearest competitor.
\begin{align}
\forall_{k \neq c}:~p_c(\x) &> p_k(\x) \label{eq:softdeep-decision}\\
\text{with} \quad p_c(\x) &= \frac{\exp(\centroid_c^\top \ba)}{\sum_k \exp(\centroid_k^\top \ba)} \quad \text{and} \quad \ba = \Psi(\x) \nonumber
\end{align}
\begin{proposition}
The decision function of Eq.\ \eqref{eq:softdeep-decision} can be expressed by the neural network:
\begin{boxnetwork}[Neuralized deep soft clustering (relative)]
\begin{align*}
\ba &= \Psi_L \circ \dots \circ \Psi_1(\x) & (\mathrm{layers}~1\dots L)\\
h_k &= \w_k^\top \ba & (\mathrm{layer}~L+1)\\
f_c &= \min_{k \neq c}\{h_k\} & (\mathrm{layer}~L+2)
\end{align*}
\end{boxnetwork}
where $\w_k = \centroid_c - \centroid_k$, and testing for $f_c \geq 0$. Furthermore, $f_c$ has a probabilistic interpretation as the log-likelihood ratio $\log (p_c(\x) / \max_{k\neq c} \{p_k(\x)\})$.
\label{proposition:softdeep-neuralization}
\end{proposition}
A proof is given in Appendix E of the Supplement. The solutions in \cite{DBLP:conf/icml/XieGF16,DBLP:conf/iconip/GuoLZY17} are also based on a soft-assignment model, where the exponential terms are replaced by t-Student distributions. The latter do not allow for a similar neural network reformulation as above, however, they still converge to hard k-means when the clusters become increasingly distant.

\smallskip

Alternatively, one can be interested in why an assignment onto a cluster exceeds a particular probability threshold. Specifically, we would like to explain the decision function:
\begin{align}
p_c(\x) &> \theta\label{eq:softdeep-decision-2}
\end{align}
where the probability scores are defined in the same way as in Eq.\ \eqref{eq:softdeep-decision}, and where $\theta$ is some value between $0$ and $1$.
\begin{proposition}
The decision function of Eq.\ \eqref{eq:softdeep-decision-2} can be expressed by the neural network:
\begin{boxnetwork}[Neuralized deep soft clustering (absolute)]
\begin{align*}
\ba &= \Psi_L \circ \dots \circ \Psi_1(\x) & (\mathrm{layers}~1\dots L)\\
h_k &= \w_k^\top \ba + b_k & (\mathrm{layer}~L+1)\\
f_c &= {\LME_{k\neq c}}^{-1}\{h_k\} & (\mathrm{layer}~L+2)
\end{align*}
\end{boxnetwork}
where $\w_k = \centroid_c - \centroid_k$, $b_k = - \log(N-1) + \log ((1-\theta)/\theta)$ and testing for $f_c \geq 0$. Furthermore $f_c = \log (p_c(\x) / (1-p_c(\x)))+\log((1-\theta)/\theta)$, i.e.\ a log-likelihood ratio plus an offset.
\label{proposition:softdeep-neuralization-2}
\end{proposition}
A proof is given in Appendix E of the Supplement. Like for the k-means and kernel k-means cases, the min-take-most propagation rule can be applied to the top layer. For the last neuralized variant featuring the $\LME$ computation, one also needs to handle the case where non-zero relevance scores $R_k$ land on deactivated neurons ($h_k=0$). To avoid this, we perform the reassignment $R_k \gets R_k \cdot (h_k / f_c)$. For further propagation of relevance scores into the neural network, we notice that all layers up to layer $L+1$ form a standard neural network. Hence, propagation rules designed in the context of neural network are applicable. For propagation rules specific to deep neural networks, we refer to the papers \cite{DBLP:series/lncs/MontavonBLSM19,DBLP:series/lncs/ArrasAWMGMHS19} which cover in particular convolutional layers and LSTM blocks.

\section{Extension to Any Clustering}
\label{section:any}

Not all clusterings can be readily obtained by \skdkmeans{} or combinations of them. Algorithms such as DBSCAN \cite{DBLP:conf/kdd/EsterKSX96}, hierarchical agglomerative clustering \cite{Gower1969}, or spectral clustering \cite{DBLP:conf/nips/MeilaS00,DBLP:conf/nips/NgJW01}, are based on a different principle, and typically lead to different cluster solutions. For these clusterings we observe however that the decision function they implement is typically based on evaluating distances between individual data points. Hence, the kernel k-means model we have proposed provides a natural surrogate for modeling the cluster assignment of these models. In particular, the identified four-layer architecture can be kept fixed, and the parameters (e.g.\ data point weightings) can be fine-tuned to fit the decision boundary. Once the model boundaries coincide, the model can be used in a second step to extract explanations. The same fine-tuning strategy can be used to handle cluster solutions that are not the sole result of a cluster algorithm but that have instead been curated by humans to match their expert knowledge.

\smallskip

Compared to a standard surrogate approach that would use a generic classifier to fit the cluster assignments, using a \skdkmeans{} surrogate ensures that the needed adjustment is minimal, thereby preventing the decision strategy of the two models to become substantially different. In particular, one minimizes the risk of introducing a Clever Hans effect into the surrogate model (cf.\ \cite{lapuschkin-ncomm19}), or removing such Clever Hans effect. The risk would indeed be that the surrogate model yields a false interpretation (too optimistic or too pessimistic) of the original model's decision strategy.

\section{Applications}
\label{section:experiments}

We have proposed to extend Explainable AI to clustering, and have contributed the neuralization-propagation technique (NEON) to efficiently extract these explanations. In the following, we demonstrate on three showcase examples how one benefits in multiple ways from enriching cluster assignments with explanations.

\subsection{Better Validation of a Clustering Model}
\label{section:text-explanation}

\begin{figure*}
\centering
\includegraphics[width=.95\textwidth,]{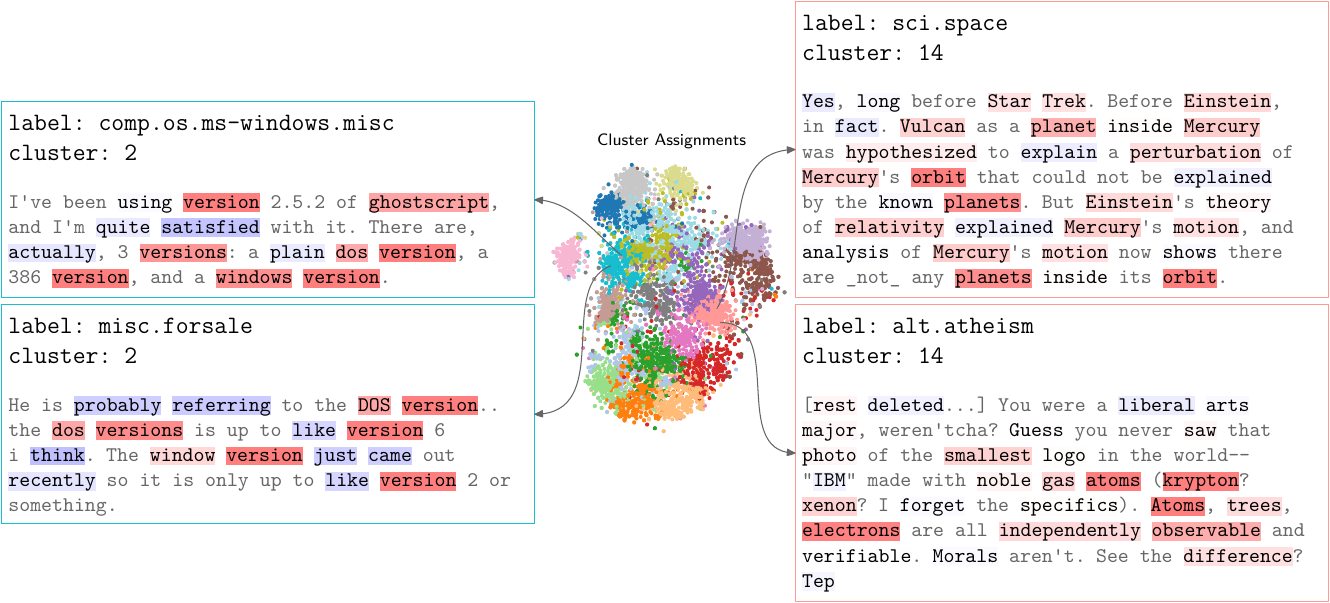}
\caption{Application of \acronym{} to the clustering of newsgroup data.  Newsgroup texts where words relevant for cluster membership are highlighted. Gray words are out of vocabulary.}
\label{fig:kernel_20newsgroups}
\end{figure*}

The following showcase demonstrates how an explanation of cluster assignments can serve to produce a rich and nuanced assessment of cluster quality that goes beyond conventional metrics such as cluster purity.

We consider for this experiment the 20newsgroups dataset \cite{DBLP:conf/icml/Joachims97} that contains messages from 20 public mailing lists, recorded around the year 1996. Headers, footers and quotes are removed from the messages. Each document $\mathcal{D}$ is represented as a collection of words defined as any sequence of letters of length at least three. Stop words are removed. Document vectors are then produced by mapping each word $t$ it contains to its tok2vec\footnote{We use spaCy {\tt md} word embeddings: \url{https://spacy.io}} representation $\varphi(t)$ (similar to word2vec \cite{DBLP:journals/corr/abs-1301-3781}), and computing the average $\x = \frac{1}{|\mathcal{D}|} \sum_{t \in \mathcal{D}} \varphi(t)$. We cluster the data using a kernel \kmeans{} model with $10$ support vectors per cluster. Initializing the kernel clustering with ground truth labels and training the kernel \kmeans{} model with an EM-style procedure (see Appendix F of the Supplement for details), the cluster assignment converges to a local optimum with the final assignment visualized in Fig.\ \ref{fig:kernel_20newsgroups} (middle).

We now focus on assessing the quality of the learned clusters. The Adjusted Rand Index (ARI) metric gives a score of 32\%, whereas the same model trained with fixed assignments to the true labels reaches 45\%. From this score, one could conclude that the algorithm has learned  `bad' clusters. 
Instead, cluster explanations, which expose to the user {\em what} in a given document is relevant for its membership to a certain cluster, will give a quite different picture. We first note that a direct application of the NEON method we have proposed to obtain such explanation would result in an explanation in terms of the dimensions of the input vector $\x$, which is not interpretable by a human as word and document embeddings are usually abstract. A more interpretable word-level explanation can be achieved, by observing that the mapping from words to document (an averaging of word vectors) and the first layer of the neuralized kernel \kmeans{}, are both linear. Thus, they can be combined into a single `big' linear layer that takes as input each word distinctly. These scores can then be pooled over word dimensions \cite{Arras2017}, leading to a single relevance score $R_t$ for each individual word $t$. These explanations can be rendered as highlighted text.

We select a few messages that we show in Fig.\ \ref{fig:kernel_20newsgroups}. The two messages on the left are assigned to the same cluster but were posted to different newsgroups (i.e.\ have different labels, and thus hurt the ARI). Here, \acronym{} highlights in both documents the term ``version''. Closely related terms like ``DOS'', ``windows'' and ``ghostscript'' are highlighted as well. The fact that ``version'' was found in both messages and that other related words were present constitutes an explanation and justification for these two messages being assigned to the same cluster.

As a second example, consider messages on the right in Figure \ref{fig:kernel_20newsgroups}, posted on two different groups, but that are assigned to the same cluster. The top message is discussing specifics of Mercury's motion, whilst the bottom message draws an analogy between physical objects and morals. The most relevant terms are related to physics, such as ``Einstein'' or ``atoms''. Also more broadly used terms (that may appear in other clusters too) like ``motion'' or ``smallest'' provide evidence for cluster membership. Here again, the words that have been selected hint at meaningful similarity between these two messages, thus justifying the assignment of these messages to the same cluster.

Overall, in this showcase experiment, minimizing the clustering objective has led to a rather low ARI. According to common validation procedures, this would constitute a reason for rejection. Instead, the cluster membership explanations produced by \acronym{} could pinpoint to the user meaningful cluster membership decisions that speak in favor of the learned cluster structure.

\subsection{Getting Insights into Neural Network Representations}
\label{section:insights-representations}

\begin{figure*}[t]
\centering
\includegraphics[width=0.96\linewidth]{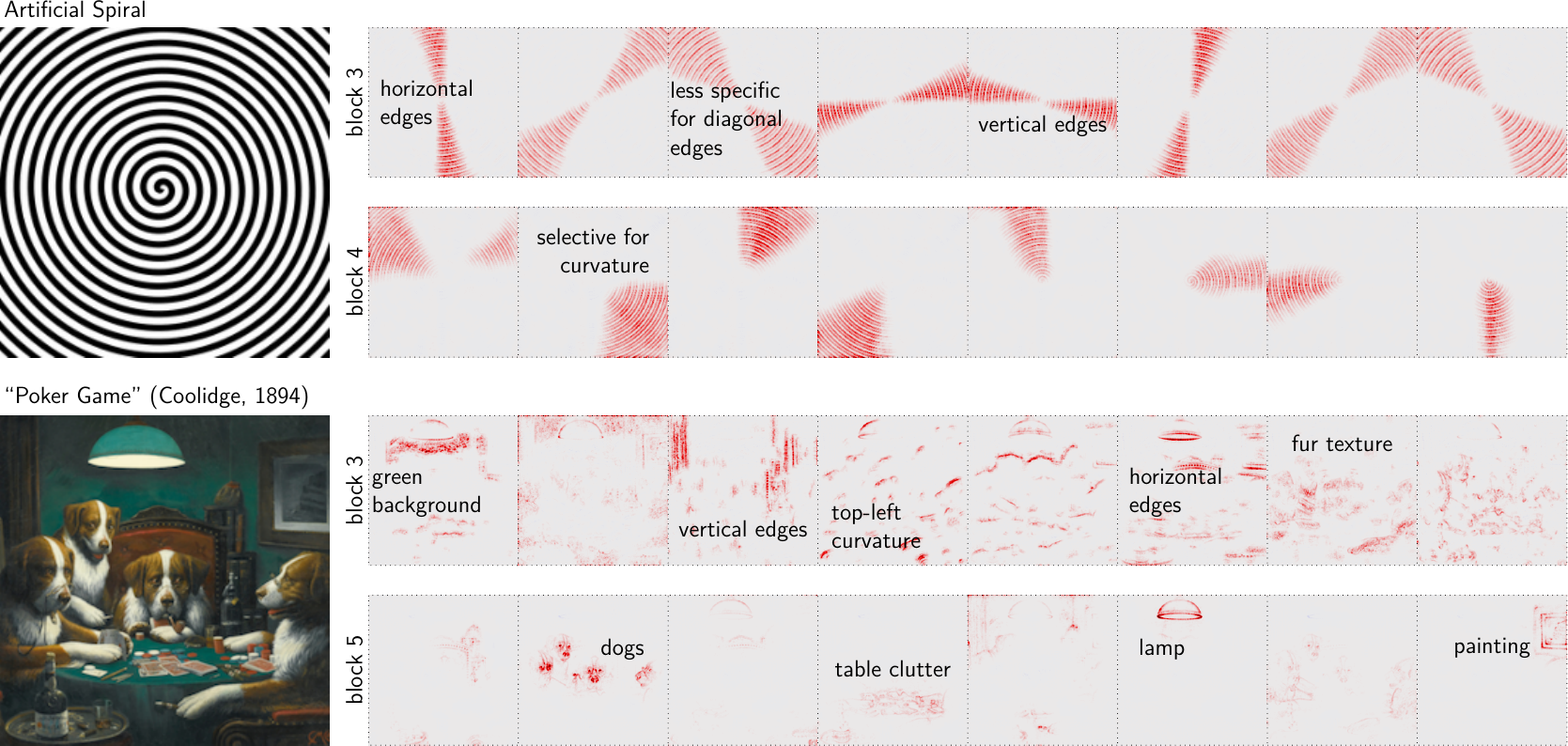}
\caption{\acronym{} analysis of images represented at different layers of a deep neural network (pretrained VGG-16). K-means clustering with $K=8$ is performed at the output of these two blocks. Each column shows the pixel-wise contributions for one of these clusters.}
\label{figure:bigimg}
\end{figure*}

Our second showcase example demonstrates how cluster explanations can be applied beyond clusters assessment, in particular, how it can be used as a way of getting insights into some given data representation $\Psi$, e.g.\ some layer of a neural network. An direct inspection of the multiple neurons composing the neural network layer is generally unfeasible as there are many such neurons, and their relation to the input is highly nonlinear. The problem of understanding deep representations has received significant attention in recent years \cite{DBLP:conf/nips/NguyenDYBC16, zhou18, lapuschkin-ncomm19}.

We consider the data representations built by the well-known VGG-16 convolutional network \cite{simonyan_very_2015}. The VGG-16 network consists of a classifier built on a feature extractor. The feature extractor is composed of five blocks alternating multiple convolutions and ReLU activations. Each block terminates with a 2$\,\times\,$2 spatial pooling, thereby creating increasingly more abstract and spatially invariant representations.

To analyze representations produced by VGG-16, we feed some image of interest into the network, leading to spatial activation maps at the output of each block. Collecting the activations at the output of a given block, we build a dataset, where each spatial location in the block corresponds to one data point. After this, we apply \kmeans{} with $K=8$ on these data points (rescaled to unit norm) and neuralize the model. For each cluster, we consider the model outputs $f_c$, and propagate these outputs backward through the network using LRP in the neuralized model and further down into the VGG-16 layers to form a collection of pixel-wise heatmaps associated to each cluster. When computing the explanations, we set $\beta$ according to our heuristic in Eq.\ \eqref{eq:heuristic}, and in convolution layers, we use LRP-$\gamma$ \cite{DBLP:series/lncs/MontavonBLSM19} with $\gamma = 0.25$ in blocks 1--3 and $\gamma = 0.1$ in blocks 4--5.

Cluster explanations are shown in Fig.\ \ref{figure:bigimg} for an artificial spiral image, and one of the well-known ``dogs playing poker'' images, titled ``Poker Game'' by Cassius Marcellus Coolidge, 1894. Images were fed to the network at resolution 448$\,\times\,$448. In the {\em artificial spiral image}, clusters at the output of block 3 map to edges with certain angle orientations as well as colors (black and white) or edge types (black-to-white, or white-to-black). Interestingly, strictly vertical and strictly horizontal edges fall in clusters with very high angle specificity, whereas edges with other angles fall into broader clusters. When building clusters at block 4, color and edge information become less prominent. Clusters are now very selective for the angle of the curvature, something needed to represent higher-level concepts. Hence, this analysis reveals to the user a specific property of the VGG-16 neural network which is the progressive building of curvature in deep representations. In the {\em Poker Game} image, we observe at block 3 a cluster that spans the green texture in the background, one that spans the fur texture associated to the dogs, and further clusters that react to edges of various orientations. After block 5, the clusters once again form higher-level concepts. There is a cluster for the big lamp at the top of the image, a cluster for the painting in the upper right, and a cluster that represents the dogs. Note that it only represents the most discriminative part of the dog, and build invariance w.r.t.\ other parts of the dogs, in particular, the fur texture. This reveals to the user how VGG-16 progressively builds high-level abstractions and become invariant to certain visual features.

To summarize, our cluster explanations could extract useful insight about the way VGG-16 represents its input from a small selection of images. In particular, our analysis does away with the high dimensionality of neural network representation by providing an explanation that fits in only $8$ heatmaps, hence easily interpretable by the user.

\subsection{Getting Insights into the Data}
\label{section:insights-data}

While Explainable AI techniques have shown helpful to shed light into the decision strategy associated to specific models and data representations, it also provides a useful tool to extract insight into the data distribution itself (exploratory data analysis). This is often desirable in scientific applications \cite{Ciriello2013,Horst2019}, where the model serves to discover interesting correlations in the data rather than being of interest on its own. Our last showcase demonstrates that NEON, in conjunction with a well-functioning clustering model, can extract such insight into the data. In particular, we find that clusters of the data can be linked to contiguous patterns in pixel space, often corresponding to the image segments provided by the user.

\begin{figure}[h!]
\centering
\includegraphics[width=.95\linewidth]{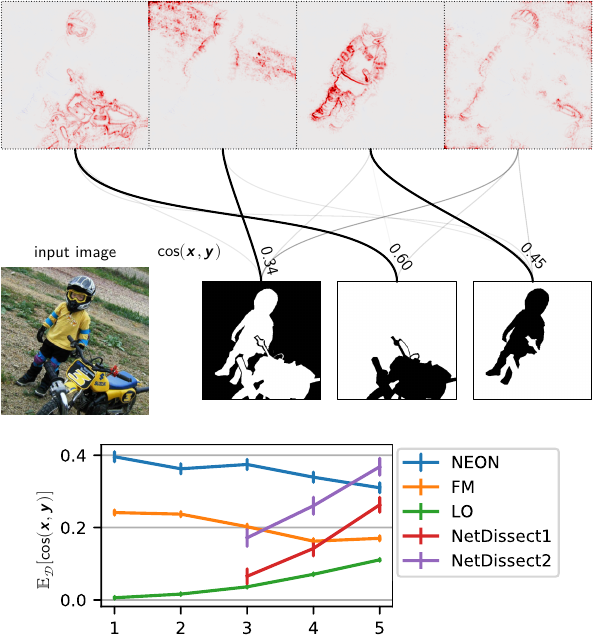}
\caption{Quantitative evaluation of \acronym{}'s ability to extract meaningful summaries. {\em Top:} The cluster explanation is matched with ground truth object segmentation masks by means of cosine similarity. {\em Bottom:} Comparison of \acronym{} to other methods. For each method we show the average cosine score over the whole dataset. Results are shown for different blocks on the $x$-axis.}\label{fig:vgg_results}
\end{figure}

To demonstrate this property of the data, we consider the PASCAL\,VOC\,2007 dataset \cite{pascal-voc-2007} which comes with segmentation masks separating the different objects. We consider a similar setting to Section \ref{section:insights-representations}, where we build a collection of $K$-cluster models based on activation vectors at different spatial locations and at a given layer of the pretrained VGG-16 network. The assignment of these activation vectors onto the learned clusters is then attributed to the input pixels using our NEON explanation framework to form a collection of $K$ heatmaps. Fig.\ \ref{fig:vgg_results} (top) shows an example of heatmaps we get for an image of a kid with a small motorbike. We observe that the attribution of cluster membership onto pixels highlights that each cluster represents distinct objects in the image, here, the kid, the motorbike and the background. 
We perform an experiment where we measure to what degree explained clusters match the different segmentation masks. Similarity between heatmaps and segmentation masks is measured by a maximum weight matching (Hungarian algorithm) between masks and clusters, where the weight is given by their cosine similarity. The procedure is depicted in Fig.\ \ref{fig:vgg_results} (middle). The matching is reduced to a single score $S\in[-1,+1]$ by averaging the cosine similarity of all matchings. A perfect score of $S=1$ can only be achieved if the clusters are strictly equivalent to the matched segmentation masks.

For comparison, we construct two simple baselines that do not make use of clustering: The first baseline takes the top-$k$ most activated (in the $L_\infty$ sense) feature maps (FM). The second baseline takes the top-$k$ most activated locations (LO). In addition we consider a recently proposed method, NetDissect \cite{zhou18}, which identifies meaningful segments of an image by thresholding spatial activation maps. Thresholds applied by NetDissect are learned in a supervised manner to match a rich set of concepts (e.g.\ {\em wood}, {\em red} or {\em carpet}) from the Broden dataset. The NetDissect1 baseline takes the top-$K$ segmentation maps. NetDissect2 takes $K$ centroids from all segmentation maps. For every method in our benchmark, we fix $K = 4$ (the average number of objects in the dataset) and apply the same LRP propagation rules for NEON, FM and LO. Examples of heatmaps produced by each method are given in Appendix J.

Average cosine similarities for each method applied at the output of each block\footnote{NetDissect only has results for blocks 3--5 due to its high computational cost in the lower layers.} are given in Fig.\ \ref{fig:vgg_results} (bottom). The \acronym{} approach clearly and consistently delivers the best results except for block 5, where NetDissect2 shows a better performance. Interestingly, the highest correlation is found in lower layers, confirming that low-level features such as color or textures are good descriptors of the spatial occupancy of an object, whereas higher-level features may build too much invariance to comprehensively highlight segments (see also Section \ref{section:insights-representations}). The higher performance of NetDissect in higher-layer can be attributed to the smoother way it renders explanation in pixel-space (cf.\ Appendix J in the Supplement), thereby `undoing' some of the invariances the neural network might have built.

\medskip

Overall, our \acronym{} approach allows to shed light into the statistics of complex data distributions, for example, by finding that clusters in image data, especially those coding for low-level information content such as texture or color, substantially correlate with image segments.

\section{Evaluation}
\label{section:evaluation}

While the section above has demonstrated the multiple practical benefits one can get from bringing Explainable AI to clustering, we would like to study here more specifically the technical ability of NEON as an explanation method for clustering. We consider a broad spectrum of desiderata of an explanation method, and evaluate NEON against a number of simple contributed baselines. We stress that the baselines we use were originally proposed for explaining classification, however, with some adaptations that we propose, they can be extended to the clustering case and therefore serve as baselines in our evaluation.

In particular, we consider integrated gradients (IG) \cite{DBLP:conf/icml/SundararajanTY17} where the explanation scores are computed by integrating the model output between the origin and the data point $\x$ following some linear path. We then apply Prediction Difference Analysis (PDA) \cite{DBLP:conf/iclr/ZintgrafCAW17,Covert2020} where we score the different dimension based on the effect on the decision function of removing the corresponding feature. The missing feature is either set to zero (PDA$_0$) or imputed using a KDE conditional sampler (PDA$_\text{cs}$), which we describe in Appendix G of the Supplement. Finally, we include four simple baselines: random attribution, squared features $\x^2$, sensitivity analysis $(\nabla f)^2$, which computes the square of the derivative along each input dimension, and a method specific to standard k-means, `nearest centroid analysis' (NCA) that computes $(\x-\centroid_k)^2-(\x-\centroid_c)^2$ where $\centroid_c$ and $\centroid_k$ are the centroids of the assigned cluster and nearest competing cluster respectively, and where the squaring operation applies element-wise.

\subsection{Desiderata and Evaluation Metrics}

In the context of explaining image classifiers, \cite{samek2017evaluating} proposed the `pixel-flipping' technique for evaluating explanations. The technique consists of constructing a plot that keeps track of decision function (in our case, this will be the cluster indicator function $g_c(\x) = 1_{\{\x \to \text{cluster}~c\}}$) as we add or remove features by order of relevance according to the explanation, and measuring the area under the curve (AUC). We start from this algorithm and adapt it to our setting. In particular, instead of flipping pixels, we consider general features, and similar to \cite{schnake2021higher} start from an `empty' data point, and add the features from most to least relevant. Missing features are inpainted using a conditional sampler built on the simple kernel density estimation (KDE) model, the details of which we provide in Appendix G in the Supplement, or replaced by zero when the input features are activations of a deep neural network. The procedure for computing the AUC is detailed in Algorithm \ref{algorithm:auc}, where the AUC output is a number between 0 and 100. The higher the AUC, the better the explanation. The analysis can be extended to a whole dataset by computing by averaging the AUC obtained for each individual data point, and repeating the whole procedure multiple times to reduce the variance produced by the KDE sampling.

\begin{algorithm}
\caption{Area under the curve (AUC) computation for a data point $\boldsymbol{z} \in \mathbb{R}^d$ and the explanation $(R_i)_i \in \mathbb{R}^d$ of its prediction.}
\label{algorithm:auc}
\begin{algorithmic}
 \STATE {$\mathcal{I} = \varnothing$}
 \STATE {$\texttt{curve} = [~]$}
 
 \vskip 1mm
 
 \FOR {\texttt{$\iota \in\,\,$argsort($(-R_i)_i$)}}
 \STATE {$\mathcal{I} = \mathcal{I} \cup \{\iota\}$}
 \STATE {$\x \sim p_\text{KDE}(\x\mid\boldsymbol{z}_\mathcal{I})$}
 \STATE {\texttt{curve.append(}$g_c(\boldsymbol{x})$\texttt{)}}
 \ENDFOR
 
 \vskip 1mm
 
 \RETURN {\texttt{area\_under(curve)}$\cdot\,100~\slash~d$}
 \end{algorithmic}
\end{algorithm}

\smallskip

Consider now the five desiderata of an explanation listed in \cite{swartout93}, namely, \textit{fidelity}, \textit{understandability}, \textit{sufficiency}, \textit{low construction overhead}, and \textit{runtime efficiency}. We argue that Algorithm \ref{algorithm:auc} captures to a reasonable extent the first three of them: \mbox{\textit{Fidelity} (\textbf{D1})}:~Algorithm \ref{algorithm:auc} keeps track of the model output as we add features. This favors techniques that explain the model output rather than some other function. \mbox{\textit{Understandability} (\textbf{D2})}:~It is desirable that the explanation is understandable by its user, e.g.\ expressible in terms of input features, and simple enough (e.g.\ a few relevant features). Algorithm $1$ implements such desiderata by verifying whether the few most relevant features returned by the explanation produce a substantial increase of the model output. \mbox{\textit{Sufficiency} (\textbf{D3})}:~The explanation should be sufficient for its user, i.e.\ provide sufficient information about the model's decision strategy. Algorithm \ref{algorithm:auc} requests a score for each individual feature (or at least a full ranking of those features). This favors explanations with this level of resolution compared to more coarse-grained explanations.

To assess the fulfilment of the last two desiderata, we proceed as follows: \mbox{\textit{Low construction overhead} (\textbf{D4})}:~The explanation technique should not be too complex or costly to implement. Our evaluation will rank explanation methods depending on whether they only need access to the decision function, access to some differentiable function reproducing the decision function, or access to the neural network internals of that function. \mbox{\textit{Runtime efficiency} (\textbf{D5})}:~The explanation should be computable quickly. In our evaluation, we will provide the algorithmic complexity of each explanation method and perform additional runtime comparisons.

\subsection{AUC Evaluation Results}

To test desiderata \textbf{D1}--\textbf{D3}, we first perform the AUC evaluation presented in Algorithm \ref{algorithm:auc} on a set of models trained on different datasets of various dimensionality and complexity. We consider first a set of standard \kmeans{} models trained on a number of datasets from the UCI repository (details and links to the datasets are provided in Appendix H of the Supplement), and where the number of clusters $K$ is determined using the elbow method \cite{kneedle}. Then, we consider more complex kernel \kmeans{} models which we train on further datasets from the UCI repository. We also consider the kernel \kmeans{} model trained on the 20newsgroup dataset \cite{DBLP:conf/icml/Joachims97} (\texttt{news} in Table \ref{tab:pixelflipping}) which we have showcased in Section \ref{section:text-explanation}. The training algorithm we have used for kernel \kmeans{} is detailed in Appendix F in the Supplement. Finally, we consider deep \kmeans{} models built on the popular STL-10 \cite{DBLP:journals/jmlr/CoatesNL11} image recognition dataset. We consider either a standard \kmeans{} model built on the features at the output of block 5 of the VGG-16 deep neural network pretrained on ImageNet (VGG-s), or the same VGG-16 network without supervised pretraining (VGG-u) and coupled with the recently proposed SCAN \cite{DBLP:conf/eccv/GansbekeVGPG20} clustering model\footnote{We train exactly the same model as in \cite{DBLP:conf/eccv/GansbekeVGPG20}, but replace the resnet-18 feature extractor by a VGG-16 feature extractor, which comes with extensively tested LRP rules \cite{DBLP:series/lncs/MontavonBLSM19,Eberle2020}. Our trained model reaches a clustering accuracy of 72.6 (compared with 76.7 for the original model \cite{DBLP:conf/eccv/GansbekeVGPG20}, but well above earlier deep clustering proposals).} for deep clustering. For each dataset and model, we set the NEON hyperparameter according to the heuristic in Eq.\ \eqref{eq:heuristic}. For deep models, we choose $\beta$ in the same way and furthermore choose the LRP rule LRP-$\gamma$ \cite{DBLP:series/lncs/MontavonBLSM19}, with the parameter $\gamma$ set heuristically to $0.1$. For these two deep clustering models, we consider as unit of interpretability the 256 feature maps at the output of block 3 of the VGG-16 network, and thus produce explanations in $\mathbb{R}^{256}$. Results are shown in Table \ref{tab:pixelflipping}.

\begin{table*}[t]
    \centering
    \caption{AUC score computed with Algorithm \ref{algorithm:auc} and serving as a proxy for the fulfillment of desiderata \textbf{D1}--\textbf{D3}. The higher the AUC score the better the explanations. We find that the proposed NEON method scores the highest for the vast majority of clustering models. Entries where methods are inapplicable or computationally prohibitive are denoted by `---'.
    }
    \begin{tabular}{crrrcrrrrrrrr}
\toprule
\multicolumn{3}{c|}{dataset} & \multicolumn{2}{c|}{model} & \multicolumn{8}{c}{methods} \\
name & \multicolumn{1}{c}{$N$} & \multicolumn{1}{c|}{$D$} & \multicolumn{1}{c}{$K$} & \multicolumn{1}{c|}{type} & \multicolumn{1}{c}{random} & \multicolumn{1}{c}{$\boldsymbol{x}^2$} & \multicolumn{1}{c}{PDA$_{0}$} & \multicolumn{1}{c}{PDA$_{\mathrm{cs}}$} & \multicolumn{1}{c}{$(\nabla f_c)^2$} & \multicolumn{1}{c}{IG-10} & \multicolumn{1}{c}{NCA} & \multicolumn{1}{c}{NEON} \\
\midrule
{\tt buddy} & 249 & 6 & 7 & kmeans & 71.77 & 75.00 & 70.06 & 70.81 & 73.32 & 74.76 & 76.41 & {\bf 78.42} \\
{\tt c2000} & 2000 & 68 & 7 & kmeans & 90.71 & 95.12 & 90.67 & 90.87 & 92.01 & 93.82 & 92.15 & {\bf 95.21} \\
{\tt hepac} & 615 & 11 & 8 & kmeans & 61.16 & 74.04 & 59.67 & 60.03 & 76.50 & 77.56 & 76.20 & {\bf 80.19} \\
{\tt seeds} & 210 & 7 & 6 & kmeans & 75.73 & 78.64 & 76.07 & 76.09 & 81.62 & 79.54 & 81.16 & {\bf 82.81} \\
{\tt winer} & 178 & 13 & 6 & kmeans & 78.36 & 85.04 & 77.01 & 78.15 & 82.99 & 85.87 & 85.10 & {\bf 87.23} \\
\midrule
{\tt news} & 250 & 300 & 20 & kernel & 40.07 & 42.83 & 51.55 & --- & 40.40 & 40.40 & --- & {\bf 54.50} \\
{\tt trpad} & 980 & 10 & 9 & kernel & 58.68 & 71.34 & 58.87 & 58.32 & 68.76 & 71.44 & --- & {\bf 74.92} \\
{\tt sales} & 811 & 52 & 6 & kernel & 82.30 & {\bf 87.28} & 82.66 & 82.33 & 86.72 & 86.51 & --- & 87.21 \\
{\tt water} & 527 & 38 & 5 & kernel & 79.30 & 87.30 & 79.13 & 78.89 & 85.01 & 86.82 & --- & {\bf 87.66} \\
{\tt whlsl} & 440 & 6 & 8 & kernel & 54.98 & 63.91 & 54.88 & 55.11 & 64.53 & 64.93 & --- & {\bf 67.37} \\\midrule
{\tt STL-10} & 5000 & 256 & 10 & deep (VGG-s) & 50.52 & 66.66 & 75.30 & --- & 56.11 & 66.99 & --- & {\bf 77.93} \\
{\tt STL-10} & 5000 & 256 & 100 & deep (VGG-s) & 32.42 & 53.34 & 48.78 & --- & 39.47 & 41.85 & --- & {\bf 65.09} \\
{\tt STL-10} & 5000 & 256 & 1000 & deep (VGG-s) & 27.32 & 50.36 & 46.63 & --- & 34.99 & 38.85 & --- & {\bf 52.38} \\\midrule
{\tt STL-10} & 5000 & 256 & 10 & deep (VGG-u\,/\,SCAN) & 58.66 & 68.54 & 75.69 & --- & 59.40 & 70.84 & --- & {\bf 85.76} \\
{\tt STL-10} & 5000 & 256 & 100 & deep (VGG-u\,/\,SCAN) & 38.72 & 49.02 & 31.77 & --- & 41.73 & 25.52 & --- & {\bf 55.38} \\
{\tt STL-10} & 5000 & 256 & 1000 & deep (VGG-u\,/\,SCAN) & 22.98 & {\bf 32.45} & 9.28 & --- & 27.63 & 6.83 & --- & 23.40\\
\bottomrule
\end{tabular}

    \label{tab:pixelflipping}
\end{table*}

\begin{figure}[t]
\centering
\includegraphics[width=\linewidth]{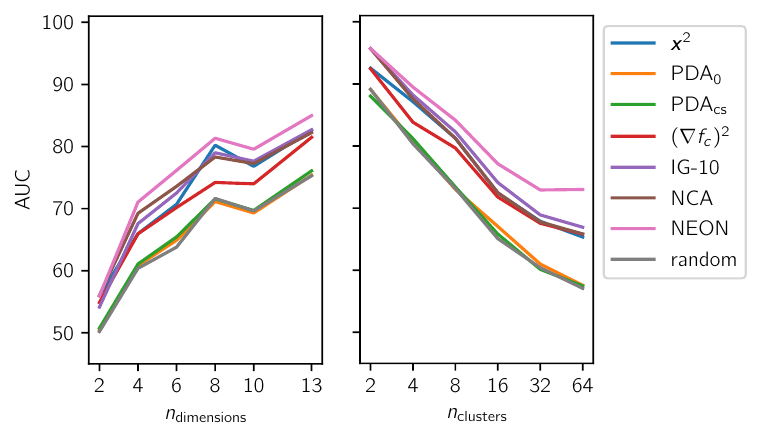}
\caption{Effect of the number of retained dimensions $d$ and the number of clusters $K$ on the AUC performance of each explanation method on the {\tt winer} dataset.}
\label{fig:varying}
\end{figure}

We observe that the proposed NEON explanation method is superior to all baselines for the vast majority of considered clustering models and datasets. We note the relatively poor performance of PDA, where the removal of individual features seems insufficient to capture the more global structure of the cluster assignment.
To get further insights into the performance of NEON, we perform an experiment where we take an existing dataset, the {\tt winer} dataset, and generate scenarios of varying complexity by training clustering model between $K=2$ to $K=64$, and also removing input features to generate dataset dimensions from $d=2$ to $d=13$. The results are shown in Fig.\ \ref{fig:varying}.

We observe that in every regime, NEON has equal or superior performance to all baselines. Anecdotally, NEON performs equivalently to NCA for $K=2$, but it start to outperform it as soon as the number of clusters grows.

\subsection{Sensitivity of NEON to Hyperparameters}

\begin{figure}[t]
\centering
\includegraphics[width=\linewidth]{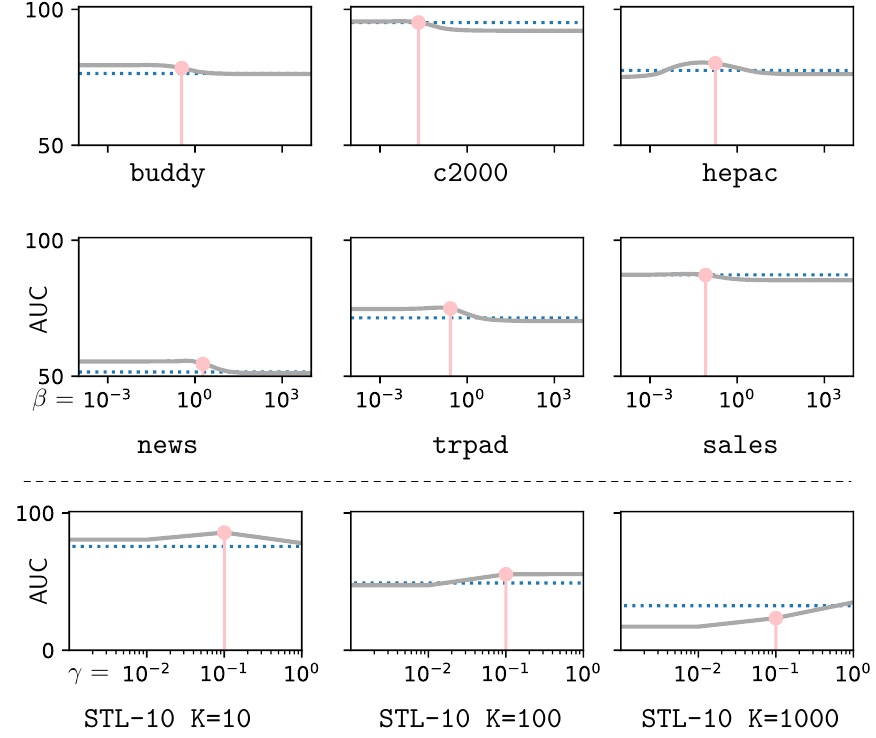}
\caption{Evaluating of NEON hyperparameters on a selection of clustering models. 1st row: \kmeans{} models, 2nd row: kernel \kmeans{} models, 3rd row: deep models (VGG-u\,/\,SCAN). The $y$-axis shows the pixel flipping AUC. The first two rows show the effect of the min-take-most parameter $\beta$, with the orange marker indicating the proposed heuristic $\beta=\mathbb{E}[f_c]^{-1}$, the dotted line is the best performing baseline (cf. Table \ref{tab:pixelflipping}). The last row shows the effect of the LRP convolution parameter $\gamma$, with the orange marker indicating our heuristic $\gamma=0.1$, and where we set $\beta=\mathbb{E}[f_c]^{-1}$.}
\label{fig:beta_plot}
\end{figure}

Unlike other baseline methods used in our benchmark, NEON comes with a `stiffness' hyperparameter $\beta$ which we have proposed to choose heuristically following Eq.\ \eqref{eq:heuristic}. For deep clustering, one also needs to choose the parameter $\gamma$ associated with the propagation in convolution layers. We would like to test the sensitivity of NEON to these parameters, first to verify the soundness of our heuristic, but also to check whether other choices of parameters lead to further improvements or conversely a degradation of NEON performance. Results are given in Fig.\ \ref{fig:beta_plot}, where we superpose on the same plot the performance at the heuristically set value for the hyperparameter (orange dot), the performance for other values of the hyperparameter (solid gray line), and the performance of best performing baseline (dotted blue line).

We observe that the simple heuristic proposed in Eq.\ \eqref{eq:heuristic} nicely correlates with the peak of AUC performance, thereby providing empirical justification for the proposed heuristic. We note that even if the hyperparameter $\beta$ is chosen inadequately, AUC performance degrades in most cases only to a minor extent. Conversely, an optimization of the NEON hyperparameters brings slight additional gains on the AUC score. Notably, the seemingly limited performance of NEON on deep clustering with $K=1000$ can be overcome by choosing a larger value for the parameter $\gamma$, in turn making NEON again the best performing method. In addition to maximizing the AUC score, the hyperparameters of NEON and the possibility to optimize them can be especially useful when bringing explainability to new tasks with specific performance metrics.

\subsection{Construction Overhead and Runtime}

Lastly, we would like to study the fulfillment by NEON of desiderata \textbf{D4} (low construction overhead) and \textbf{D5} (runtime efficiency), comparatively to other methods in our benchmark. We resort to a qualitative analysis for \textbf{D4}, where we categorize methods according to what needs to be constructed additionally to the clustering decision function. Results are shown in Table \ref{table:d4d5} (second column). The symbol `--' indicates that we do not even need the decision function, `$g_c$' indicates that we need the decision function only, `$\nabla f_c$' indicates that we need a differentiable surrogate function $f_c$ and its gradient, `$(\centroid_c)_c$' indicates that we need the cluster centroids, and finally, `NN' indicates that we need the neural network equivalent of the surrogate function $f_c$. The proposed NEON method has the highest overhead in our benchmark as it requires a neural network equivalent. However, since we have already derived these neural network equivalents in the technical sections, there is no significant obstacle to apply NEON on the studied models (\kmeans{}, kernel \kmeans{}, deep clustering, and related).

\begin{table}[t]
\centering
\caption{Fulfillment of low construction overhead and runtime efficiency desiderata for the methods in our benchmark.}
\begin{tabular}{l|c|cc}\toprule
Method & Overhead (\textbf{D4}) & \multicolumn{2}{c}{Runtime (\textbf{D5})}\\
&& standard & kernel\\\midrule
$(\x - \widetilde{\x})^2$ & -- & $\mathcal{O}(d)$ & $\mathcal{O}(d)$\\[1mm]
PDA & $h_c$ & $\mathcal{O}(Kd^2)$ & $\mathcal{O}(Kd^2p)$ \\[1mm]
$(\nabla f_c)^2$ & $\nabla f_c$ & $\mathcal{O}(Kd)$ & $\mathcal{O}(Kdp)$ \\[1mm]
IG-10 & $\nabla f_c$ & $\mathcal{O}(10 Kd)$ & $\mathcal{O}(10 Kdp)$\\[1mm]
NCA & $(\centroid_c)_c$ & $\mathcal{O}(Kd)$ & ---\\[1mm]
NEON & NN & $\mathcal{O}(Kd)$ & $\mathcal{O}(Kdp^2)$ \\\bottomrule
\end{tabular}
\label{table:d4d5}
\end{table}

Regarding the runtime efficiency (\textbf{D5}), we perform a complexity analysis of the different explanation methods, where $d$ is the number of input dimensions, $K$ is the number of clusters, and $p$ is the number of support vectors per cluster in the kernel \kmeans{} case. Results are shown in Table \ref{table:d4d5} (last column). We observe that for \kmeans{}, NEON computational cost is lower or equal to most explanation methods, by only requiring a single forward and backward pass, whereas several explanation methods need to evaluate the model multiple times. (An empirical runtime comparison to all baselines for various \kmeans{} models can be found in Appendix I of the Supplement.) For kernel \kmeans{}, results are more balanced, with NEON being slower than simple sensitivity analysis, but running faster than the more advanced PDA and IG competitors if the number support vectors is smaller than the number of input dimensions or the number of integration steps respectively. Hence, while for standard k-means, we can generally claim that NEON has high efficiency, for kernel \kmeans{}, one need to additionally ensure that the number of support vectors remains small, typically less than $10$.

\medskip

Overall, we have demonstrated in our evaluation that NEON fares on average the highest, comparing favorably to all competitors when considering the multiple aspects that enter into the assessment of an explanation method. Therefore, NEON constitutes so far the most appropriate and powerful method for tackling the problem of explaining cluster assignments.

\section{Conclusion}

We have contributed by for the first time bringing Explainable AI to clustering and have proposed a general framework, called neuralization-propagation, for explaining cluster assignments of a broad range of clustering models. The proposed method converts, \textit{without retraining}, the clustering model into a \textit{functionally equivalent} neural network composed of detection and pooling layers. This conversion step which we have called `neuralization' enables cluster assignments to be efficiently attributed to input variables by means of a reverse propagation procedure.

Quantitative evaluation shows that our explanation method is capable of identifying cluster-relevant input features in a precise and systematic manner, from the simplest k-means model to some of the most recent proposals such as the SCAN deep clustering model \cite{DBLP:conf/eccv/GansbekeVGPG20}. The performance remains high across all considered data types, in particular, abstract vector data, text, natural images, or neuron activations.

The method we have proposed complements standard cluster validation techniques by providing a rich interpretable feedback into the nature of the clusters that are built. Furthermore, when paired with a well-functioning clustering algorithm, it provides a useful tool for exploratory data analysis and knowledge discovery where complex data distributions are first summarized into finitely many clusters, that are then exposed to the human in an interpretable manner.

\section*{Acknowledgements}

This work was supported by the German Ministry for Education and Research under Grant Nos.
01IS14013A-E, 01GQ1115, 01GQ0850, 01IS18025A,
031L0207D, and 01IS18037A, and the German Research Foundation (DFG) in the DAEDALUS graduate school and as Math+:
Berlin Mathematics Research Center (EXC 2046/1, project-ID:
390685689). KRM was partly supported by the Institute for Information \& Communications Technology Planning \& Evaluation (IITP) grant funded by the Korea government (No. 2017-0-00451, No. 2017-0-01779).

\bibliographystyle{IEEEtran}
\bibliography{paper}

\end{document}